# Thematic Working Group 5 - Artificial Intelligence (AI) literacy for teaching and learning: design and implementation

*Mary Webb - King's College London (TWG co-leader)*
*Matt Bower - Macquarie University (TWG co-leader)*
*Ana Amélia Carvalho -University of Coimbra*
*Fredrik Mørk Røkenes - University of Oslo*
*Jodie Torrington - Macquarie University*
*Jonathan D. Cohen - Georgia State University*
*Yousra Chtouki - Al Akhawayn University in Ifrane*
*Kathryn MacCallum, University of Canterbury, NZ*
*Tanya Linden, The University of Melbourne, Australia*
*Deirdre Butler, Dublin City University*
*Juliana E. Raffagheli - University of Padova*
*Henriikka Vartiainen - University of Eastern Finland*
*Martina Ronci - Université Paris Cité*
*Peter Tiernan - Dublin City University*
*David M. Smith - Purdue University Global*
*Chris Shelton - University of Chichester*
*Joyce Malyn-Smith - Education Development Center*
*Pierre Gorissen - HAN University of Applied Sciences*

TWG 5 focused on developing and implementing effective strategies for enhancing AI literacy and agency of teachers, equipping them with the knowledge and skills necessary to integrate AI into their teaching practices. Explorations covered curriculum design, professional development programs, practical classroom applications, and policy guidelines aiming to empower educators to confidently utilize AI tools and foster a deeper understanding of AI concepts among students.

## Introduction

OpenAI's release of ChatGPT3 in November 2022 marked a watershed moment in AI's evolution, garnering significant public interest and media coverage. ChatGPT and other generative AI (GenAI) systems demonstrated the ability to generate extended text responses to diverse natural language prompts, often mimicking intelligent human production. Educational applications of non-generative or 'analytical' AI had previously been developing, including bespoke learning platforms, adaptive assessment systems, intelligent predictive analytics, and conversational agents as discussed in EDUsummIT 2019 (Webb, Fluck et al. 2021). However, GenAI's capacity to produce human-like text responses to varied requests represented a significant advancement that could potentially disrupt traditional educational processes.

On the one hand, the capacity for students to use generative AI to complete tasks for them has raised particular concerns around plagiarism and academic integrity, as well as potential negative impacts on creativity, agency and critical thinking. On the other hand, the possible use of generative AI to amplify human creativity and productivity has the potential to affect individuals from every walk of life and professionals across all industries (Ghosh et al., 2025). Similarly, the use of AI in education by teachers is growing, with concerns around issues such as overreliance or bias, but also opportunities to enhance teaching, assessment, and personalized learning (Ifenthaler et al., 2024). Consequently, knowing how to effectively navigate a world with increasingly powerful generative AI and developing AI literacy has become an imperative for both students and teachers.

## Understanding AI Literacy

Building on contemporary research and policy frameworks, AI literacy can be conceptualized as:

> *the integrated set of knowledge, skills and attitudes that enables individuals to understand how AI works and affects society, use AI tools responsibly and effectively, critically evaluate AI outputs and limitations, communicate or collaborate with AI systems, and—at advanced levels—design or adapt AI solutions* (Long & Magerko, 2020; Chiu et al., 2024; European Commission & OECD, 2025).

AI literacy is not simply about learning to use AI tools and how AI works, but also an applied knowledge and understanding of when and how to use it responsibly for (and while) learning (see Figure 1). Therefore, we propose that the concept of AI literacy needs to expand to define two distinct components: 1) learning *about* AI (foundational AI knowledge and skills); and 2) learning *with* AI (practical AI integration).

- **Learning *About* AI** focuses on technology education; learning what AI is and is not from a historical, cultural, and societal perspective, learning about the distinction between explanatory AI and generative AI, and being able to explanation how AI works in ways that can be understood by a non-technical audience
- **Learning With AI** involves deliberate use and integration into learning activities; using AI as a co-collaborator/learning partner/thought partner, applying pedagogies focusing on dialogue between teacher-student, student-AI, student-student, using AI as a seamlessly integrated teaching tool, which is a part of a learning environment (e.g., adaptive learning tool, automatic speech recognition).

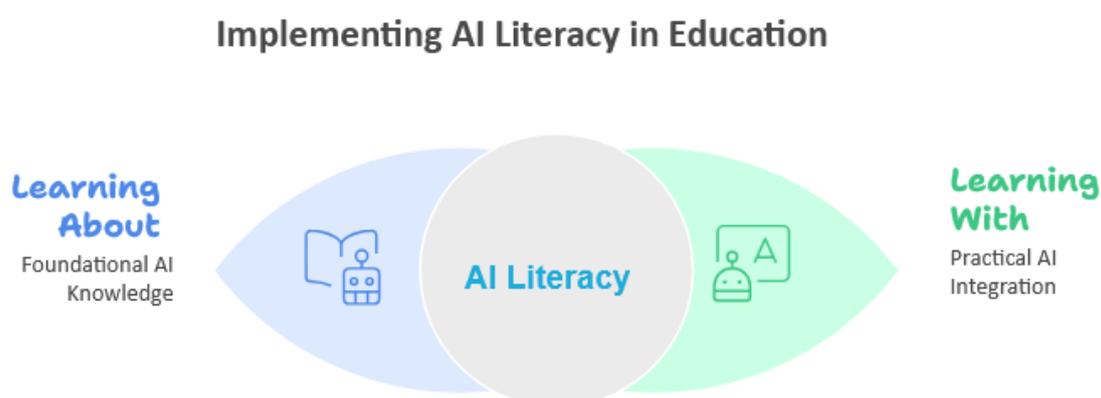

*Figure 1: Conceptualizing AI literacy in education*

AI literacy is integral to the evolving concept of **digital literacy**, reshaping how individuals access, evaluate, and produce information by influencing competencies such as media, data, and information literacy. As digital literacy expands, it increasingly demands the integration of AI and generative AI tools into everyday practices, emphasizing the need for individuals to develop skills in engaging with these technologies. AI literacy encompasses digital literacy and competence, critical awareness, and ethical understanding, and extends much further than technical tool proficiency. As AI technologies become embedded within the entire educational landscape, we are compelled to reconsider not only what we teach, but how we conceptualize teaching and learning itself.

## Developing Student AI Literacy

A number of supporting frameworks and resources are available to support the development of AI literacy. These include:

- **EC/OECD AI Literacy Framework (2025)** — 22 AI literacy competences for school learners (EC & OECD, 2025)
- **UNESCO AI Competency Frameworks (2024)** — parallel matrices for students and teachers describing knowledge, skills and values for responsible AI engagement (Miao, & Cukurova, 2024; Miao, Shiohira & Lao, 2024)
- **AI4K12 "Five Big Ideas"** — a conceptual spine for K-12 curricula, with grade-band progressions linking AI concepts to practices (AAAI & CSTA, 2025)
- **DigComp 2.2 (EU)** — embeds AI under "Digital Content Creation" and "Problem Solving," illustrating links between AI, data and media literacies (Vuorikari et al., 2022)

AI literacy frameworks worldwide share a number of common concepts: **Technical AI knowledge** (understanding AI concepts, data, algorithms) and **the ability to use AI tools and applications** are universally emphasized. A **human-centered perspective** is common: frameworks encourage critical thinking about AI's role and promote uniquely human skills (creativity, empathy, judgment) that complement AI. Each also stresses **ethics and societal impact** — teaching students to consider fairness, transparency, and the consequences of AI on people. Across these models a learning continuum is common, from recognition to use to evaluation to design and creation. As society becomes more AI literate we anticipate frameworks for AI Fluency will emerge describing in more detail the trajectory from AI user to AI developer/creator.

While research evidence on the best ways to develop AI literacies is still equivocal, there are a number of emerging principles that can be used to guide teaching *about* AI and teaching *with* AI.

When teaching *about* foundational AI knowledge, educators may choose to:

- **Centre the role of humans in creating AI** and avoid anthropomorphisation
- **Develop understanding about the differences between types of AI** and their affordances and constraints (e.g., predictive/explanatory AI vs. GenAI, Intelligent Tutoring Systems, Learning Analytics)
- **Emphasize affordances of AI technologies, rather than specific AI tools**, supporting learners' abilities to make judgements about which AI tools are most appropriate to the learning activity or pedagogical approach
- **Identify misconceptions** that may hinder understanding or use of AI and find ways to address them that are consistent with learners' technical knowledge

- **Scaffold AI learning from foundational technical, practical and ethical perspectives** to the needs of the learners and their contexts

When teaching *with* AI, where AI tools are integrated into learning activities, educators can be encouraged to:

- **Understand the specific affordances of AI tools** (i.e. capabilities and limitations), such as adaptive feedback, content generation, or personalized support, and align them with evolving pedagogical approaches;
- **Experiment with and evaluate pedagogical approaches that might be enabled by AI** such as problem-based learning, collaborative learning, AI-mediated peer feedback and real-time formative assessment;
- **Consider the learning design of activities for effective student learning,** recognizing that initial, independent struggle can be more productive for deeper learning than receiving immediate AI assistance (see Kapur, 2016);
- **Support student self-regulated learning** by explicitly focusing on developing students' self regulated learning both in individual and group learning situations through coregulation and socially shared regulation of learning (building on work from EDUsummIT 2022, see Prasse et al., 2024);
- **Include pedagogical strategies to support students to manage and monitor their effort** while using AI tools;
- **Orient learners to regard AI as a tool for deepening learning, not as an answer machine**;
- **Help learners identify what tasks may be offloaded to AI** to reduce cognitive load and what tasks require deep understanding for learning;
- **Emphasize human agency** — learners must be taught to evaluate generative AI output and to always maintain a critical/evaluative lens when engaging with AI.

While further research will be needed to determine which methods and strategies for teaching about and with AI are most effective, the principles above serve as initial guidance for educators.

## Ethical Considerations and Critical Perspectives

AI ethics and the literacies connected to it are inextricably linked to their surrounding context. AI, like any other technology, is a socio-technical product that embeds human interests, cultural perspectives, and intentions of social impact. Drawing on Cultural Historical Activity Theory (CHAT) (Engeström, 1987), we understand that AI tools mediate human activities within complex systems of relationships, community structures, and divisions of labor. This systemic view means that ethical AI literacy cannot simply involve adopting ethical standards or guidelines and ensuring compliance. Instead, it requires understanding how AI tools become embedded within specific educational contexts and how they shape—and are shaped by—local practices, values, and power dynamics.

The integration of AI in education has sparked a range of critical debates:

1. **Protecting privacy and safety**: AI in education often involves collecting vast amounts of student data, including academic performance, behavioral patterns, and biometric information, raising serious privacy and safety concerns;

2. **Bias and fairness**: AI systems can perpetuate existing social and cultural biases from their training data, leading to unfair decision-making and reinforcing systemic inequities;
3. **Trust and transparency**: AI decision-making processes are often opaque, raising questions about accountability when AI systems affect students' learning, development, and well-being;
4. **Equity in AI literacy development**: Ensuring all students can develop critical understanding of AI alongside the creative, ethical, and critical thinking skills needed for informed decision-making in a datafied society;
5. **Offloading learning**: AI misuse may prevent rather than enhance learning, particularly when used to bypass critical thinking and learning processes;
6. **Role of teachers**: Addressing concerns that AI might replace human teachers while recognizing that tailored epistemic, emotional, metacognitive and social scaffolding from teachers and peers remains essential for human learning;
7. **Environmental concerns**: training of AI models and disposal of e-waste raise sustainability issues that affect the whole planet and that can be addressed when deciding whether or not to use these tools.

These ethical considerations cannot be broached through narrow silos of thinking, and instead we draw on Bronfenbrenner's (1994) ecological framework to examine how ethics are shaped across four interconnected layers (see Figure 2).

**The Individual**: Personal interests, goals, values, and digital skills shape engagement with AI tools. However, many AI mechanisms remain invisible to users, creating challenges for ethical decision-making and raising questions of trust when interacting with opaque systems.

**School Context**: Teachers work within specific local contexts with particular AI tools, curriculum content, and diverse learners. Ethical decisions are embedded in everyday pedagogical practice as teachers decide whether, when and how to integrate AI tools whilst maintaining professional autonomy.

**State and National Context**: Educational structures such as national policies, curricula, assessment practices, and funding models shape what is taught and which technologies are available. Policy decisions may prioritize efficiency and standardization, potentially conflicting with local pedagogical needs and professional ethics.

**Cultural Values and Norms**: Broader cultural contexts encompass societal norms, value systems, and socio-technical imaginaries that shape how AI and ethics are perceived and enacted. Cultural values influence which ethical concerns are foregrounded and whose perspectives are recognized in shaping AI futures.

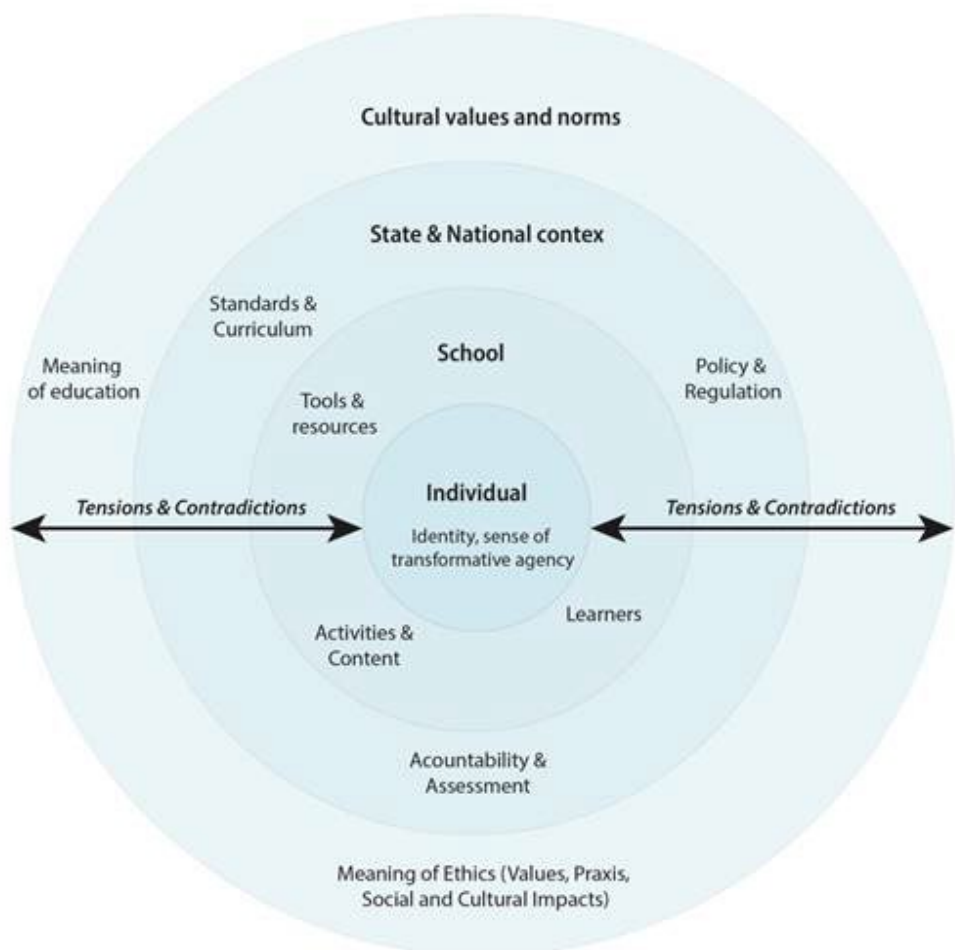

*Figure 2. An Ecological Framework for AI Ethics*

These levels are not independent but entangled, overlapping, and often contradictory. Ethical AI literacy involves developing the capacity to critically evaluate, resist when necessary, and reshape AI implementation based on educational values and local contexts. This requires maintaining human agency at the center of the learning process whilst harnessing AI's potential to transform educational challenges into opportunities for more equitable and inclusive learning. Understanding AI ethics requires recognizing that AI tools are not neutral instruments but are embedded within complex socio-cultural ecosystems. Rather than framing AI as inherently good or bad, our focus must shift to how AI is designed and integrated into learning processes.

## Implementation Opportunities and Challenges

Disruptions typically subtend a range of opportunities and challenges, and the rapid emergence of AI is no exception. Key opportunities include:

- **Enhanced human interaction**: More time for individual guidance and social-emotional support;
- **Personalized learning**: Individualized instruction and differentiated content delivery;

- **Critical thinking development**: Students learning to question AI outputs and maintain agency;
- **Systemic transformation**: Transparent, accountable technology with community oversight.

However, there are a number of challenges to address:

- **Policy-practice gaps**: Insufficient implementation planning and lack of common language across stakeholders;
- **Professional development deficits**: Limited research on AI competencies needed for individual flourishing and social advancement;
- **Static frameworks**: Frameworks that don't evolve with practice and research;
- **Human agency concerns**: Power concentration and governance in edtech industry, risk of technological dependency, and loss of decision-making.

## Key insights from other TWGs

The implications of generative AI permeate to all areas of education. However, the work of TWG5 had particular resonance with TWG3 (Tech-enabled inclusivity) and TWG1 (Bridging the gap between research and practice).

From TWG3, we learnt that when thinking about the individual, we need to remember that people are connected to family, culture, and social contexts. Hence empowering learners also means learning together by honoring their cultural heritage so that, in time, they can be empowered to become actors that are comfortable in their (digital) knowledge and can adapt their skills. This links to the ecological framework being used by TWG5 to conceptualize AI literacies and ethics. As well, TWG3 helped us to understand that learning how something works is an empowerment issue. For instance, knowing how a car works may not be needed to drive it, but can be crucial to understand when it breaks and not knowing how it works might introduce risks for the driver and others. This relates directly to TWG5's conception of learning 'about' AI.

TWG1 pointed out that we shouldn't only be concerned about a digital divide that relates to access, but also a 'digital use divide' that relates to practice. If people are provided with access but do not practice, they may still be at a social disadvantage. TWG1 also pointed out that 'Inclusion' encompasses an element of ordering and that the ideas of 'co-development' and 'co-design' may be preferable. They underlined that Indigenous and marginalized communities need to be part of discussions relating to AI literacy. This links with TWG5 points relating to accessibility and the need for AI to be considered as a human product.

## Moving Forward: Strategies and actions

## Strategies and actions for policy makers

As AI becomes increasingly pervasive and harder to disentangle from educational systems, educators must resist both the urge to block it entirely and the tendency to embrace it uncritically. Supporting learners and educators to engage ethically with AI in education demands a multi-layered and systemic perspective - one that accounts for the interplay and tensions between individual decision-making, classroom practices, institutional opportunities, and wider cultural structures.

There is an evident need for policy-makers, state authorities and local school administration to create affordances and structures that support the work and agency of teachers. Instead of top-down models, teachers and local communities should be involved in co-design and co-creation of digital ecosystems and educational activity structures, enabling them to become transformative educators.

## Strategies and actions for educators

Teachers need thoughtful, pedagogically-informed AI literacy teaching — an approach that interrogates the role of AI, preserves human values and agency, and ensures that learning remains a process of meaning-making, growth, and transformation. Teacher mindsets, beliefs and attitudes about AI shape how it is valued and enacted in classrooms. Supporting teachers to reflect on their pedagogical values and practices is crucial for creating space for AI that is equitable, inclusive, and empowering.

We must be clear about where AI belongs (and doesn't belong) in the learning process. While AI has become ubiquitous, teachers need to be explicit about how AI should or should not be used. This requires teachers to have the AI literacy to discern what is AI, how it can be managed, and how to deliberately integrate AI in a manner that supports learning rather than replacing it.

Self-regulation is a critical issue for practitioners to address in their practice. When integrating AI in the classroom, teachers need to navigate the tension between efficiency and deep engagement with educational processes. Encouraging students to move beyond passive consumption toward active construction of knowledge requires an understanding about what tasks can be offloaded to AI, and which ones require effort to be applied in order to develop deep learning (Burns et al., 2025). The shift from product to process in assessment is central here: we need to prioritize how students learn, not just what they produce.

## Strategies and actions for researchers

There is a clear need for interdisciplinary, longitudinal research on the effects and impacts of AI literacy initiatives as well as for understanding the tensions that will inevitably arise when transforming existing systems and social practices in education. Critical and ethical approaches for AI education call for cross-boundary collaboration and systematic action across different levels of the educational system.

This does not mean socializing individuals and communities into existing practices and conditions, but supporting them to become transformative agents who can critically explore and question the status quo, develop informed stances and voices, imagine alternative possibilities, and take collective action toward more just and sustainable AI futures.

## Actions from TWG5

TWG5 is a large and cohesive team that was productive in their Dublin meeting and is ambitious in their future goals. As well as this eBook chapter, the team plans the following three publications:

1. Navigating AI's Educational Future: Expert Scenarios and Implications for Teaching and Leadership

2. Revisiting Pedagogical Content Knowledge (PCK) development with Generative AI

3. Situated AI Ethics: A Cultural-Historical and Ecological Framework for Education

The team also intends to respond to emerging research and publication opportunities, based on developments in the AI in Education field.